\documentclass{iosart2c}
\usepackage[utf8]{inputenc}
\usepackage[T1]{fontenc}
\usepackage{times}
\usepackage{natbib}

\usepackage{hyperref}
\usepackage{amsmath}
\usepackage{dcolumn}
\usepackage{endnotes}
\usepackage{graphicx}

\usepackage{booktabs}
\usepackage{soul}
\usepackage{xcolor}

\pubyear{2019}
\volume{0}
\firstpage{1}
\lastpage{1}

\begin{document}
\begin{frontmatter}

%\pretitle{Pretitle}
\title{\#Brexit: Leave or Remain? The Role of User's Community and Diachronic Evolution on Stance Detection}
\runningtitle{\#Brexit: Leave or Remain?}
%\subtitle{Subtitle}

\author[A]{\fnms{Mirko} \snm{Lai}\thanks{Corresponding author. E-mail: lai@di.unito.it}}
and
\author[A]{\fnms{Viviana} \snm{Patti}}
and
\author[A]{\fnms{Giancarlo} \snm{Ruffo}}
and
\author[B]{\fnms{Paolo} \snm{Rosso}}
\runningauthor{M. Lai et al.}

\address[A]{Dipartimento di Informatica, Universit\`a degli Studi di Torino, C.so Svizzera 185, 10149, Turin, Italy}
\address[B]{PRHLT Research Center, Universitat Polit\`ecnica de Val\`encia, Camino de Vera s/n. 46022 Valencia, Spain}

\begin{abstract}
Interest has grown around the classification of stance that users assume within online debates in recent years. Stance has been usually addressed by considering users posts in isolation, while social studies highlight that social communities may contribute to influence users' opinion. Furthermore, stance should be studied in a diachronic perspective, since it could help to shed light on users' opinion shift dynamics that can be recorded during the debate.
We analyzed the political discussion in UK about the BREXIT referendum on Twitter, proposing a novel approach and annotation schema for stance detection, with the main aim of investigating the role of features related to social network community and diachronic stance evolution. Classification experiments show that such features provide very useful clues for detecting stance.
\end{abstract}


\end{frontmatter}

\section{Introduction}
Online debates are a large source of opinion-sharing dialogue on current socio-political issues, and several works rely on finer-grained sentiment analysis techniques to analyze politics. In the last decade, social media gained a very significant role in the public debate, specially in political activism. Indeed, several politicians - or their staff - actually use Twitter or other social media directly to spread their opinions or to reinforce their political campaign. On the other hand, Twitter users take part in the discussion, in particular during elections and events of public interest. Social media provide a powerful experimental tool to investigate how individuals are exposed to diverse viewpoints. Although there is an on-going scientific discussion on the existence and the real extent of the so called ``echo chambers'' and ``filter bubbles'', it is however possible to observe that online and offline forms of political participation can have both positive and negative effects \cite{Elejalde2017,Facebook-Theocharis2016}.
The ever growing number of messages posted on social media platforms has progressively motivated the increasing need of intelligent systems able to assess the contents that users generate. 
In particular, a growing interest has been expressed for the classification of users' stance, i.e. the detection of positions pro or con a particular target entity that users assume within polarized debates, applied to data from microblogging platforms such as Twitter \cite{mohammad-EtAl:2016:SemEval}.

Social studies highlight that users' social communities can play a crucial role in determining stance within polarized debates since social relations may contribute to influence users' opinion. It has been observed that individuals that share the same stance toward a specific target are likely to belong to the same community, thus it would be an interesting cue to exploit in stance detection.
Moreover, stance should be studied in a diachronic perspective, since people may change their opinions or
their communication style after some particular events that happened when the debate is still active. However, stance has been mostly addressed by considering users posts in isolation, focusing on the content analysis of the single posts, without considering the surrounding context.
In this work, we aim at investigating this issue, by proposing a new approach to stance detection (SD henceforth) where the role of three orthogonal contextual facets, which may influence user's stance, is explored: (i) social network community, (ii) diachronic evolution and (iii) common knowledge on the entities involved in the debate. 

Our approach relies on the use of content analysis techniques within a computational social science scenario \cite{lazer2009life} for extracting from the social media data information on social relations between users sharing contents and opinions towards a given target entity.

In order to evaluate our approach, we analyzed the political discussion in United Kingdom (UK) about the European Union membership referendum, held on June 23rd 2016, commonly known as BREXIT, an abbreviation for ``British exit'', collecting about 5M of English tweets containing the hashtag \#brexit. We propose a novel stance detection annotation schema to apply to our social media data that takes into account diachronic evolution, allowing to study stance from a diachronic perspective.  
As a side product of our work, we released the Twitter diachronic Brexit corpus (TW-BREXIT, herceforth), where diachronic triplets of tweets posted by 600 users active in the debate have been annotated for stance.
The core idea is to consider the evolution of the user's stance over the time monitoring her posts in different time windows corresponding to three 24-hour time steps forthcoming elections.

%Thus we decided to define a triplet as a collection of three random tweets written by the same user in a defined time interval. A triplet was collected for each user in each of the three temporal windows.
%In Lai et al. \cite{LaiExtractingGraph2017}, a first analysis of the social network of the users was carried out, here, our stance model, including the novel set of contextual features mentioned above, has been evaluated in a set of stance classification experiments on the TW-BREXIT corpus.
Overall, the analysis of the corpus and the classification experiments show that: 

\begin{itemize}
\item Analysis of the community network structure helps to improve the performance of the SD task. Therefore, the intuition that user's stance is strongly related to the social media network community, in accordance with the homophily principle \cite{mcpherson2001birds,CalaisGuerra:2011:BOT:2020408.2020438,ICWSM112847}, seems to be confirmed in our setting.
\item User's stance changes after relevant events, in accordance with theoretical studies from political sciences  \cite{gelman1993american}.
This confirms the importance of taking a diachronic perspective in the SD task.
\item Context-based features include community, diachronic evolution, and common-knowledge. All three at once obtain significant results on ablation test.

\end{itemize}

The corpus, the annotation guidelines and the source code of the experiments are available for the community to foster reproducibility of the experiments\footnote{\url{https://github.com/mirkolai/leave-or-remain}}.

The paper is structured as follows.
Related work is discussed in Section~\ref{se:related}. Section~\ref{se:features} describes the features proposed in our model. Section~\ref{se:dataset} presents the development of the TW-BREXIT corpus. Classification experiments are reported and discussed in Section~\ref{se:experiments}. Section~\ref{se:conclu} concludes the paper.

\section{Related Work}
\label{se:related}

\subsubsection*{Political sentiment and stance detection} Recent trends in monitoring political sentiment consist in considering texts from social media and in exploiting computational techniques for extracting information about the political landscape in the offline world. In particular, techniques such as sentiment analysis (SA) and opinion
mining \cite{Pang:2008:OMS:1454711.1454712}, developed within the context of computational linguistics for extracting several kinds of information about humans' behavior, can be specifically declined
for monitoring political contents and are gradually considered as especially useful
for this purpose, with possible different focuses: detecting users stance, detecting
the polarity of messages expressing opinions about candidates in political elections %and so on
\cite{BoscoPattiEncy2017},  detecting deception in text \cite{Hernandez2018impactpolarity}.
Among such new areas derived from SA, 
stance detection is one of the most interesting
\cite{Anand:2011catsanddogs,Lin2006WhichSideAreYouon}. 
In 2016 a shared task on Stance Detection on Twitter has been proposed at SemEval-2016 (SemEval2016-Task6).
Mohammad et al. \citep{mohammad-EtAl:2016:SemEval} describe the task as follows:
``Given a tweet text and a target entity (person, organization, movement, policy, etc.), automatic natural language systems must determine whether the tweeter is in favor of the target, against the given target, or whether neither inference is likely''. The SemEval2016-Task6 dataset included six commonly known targets in the United States, such as: ``Atheism'', ``Climate Change is a Real Concern'', ``Feminism Movement'', ``Hillary Clinton'', ``Legalization of Abortion'', and ``Donald Trump''. 

Most participating teams exploited standard text classification features such as n-grams and word embedding vectors. Some SA features relying on well-known lexical resources, such as EmoLex \cite{mohammad2013crowdsourcing}, MPQA \cite{Wilson:2005:RCP:1220575.1220619}, Hu\&Liu \cite{Hu:2004:MSC:1014052.1014073}, and NRC Hashtag \cite{DBLP:journals/corr/MohammadKZ13}, were also exploited. 
Baselines were established based on n-grams, char-grams and Majority Class (MC), but no team outperformed such baselines, confirming the difficulty of the task. 
%Further information about the task and %the systems can be found in %\cite{mohammad-EtAl:2016:SemEval}.
An enhanced version of the annotated corpus for SD used in SemEval2016-Task6 was subsequently released, which includes two additional labels: the opinion target, and the sentiment polarity \cite{sobhani-mohammad-kiritchenko:2016:interactionsentiment}.  
%In some recent works 
Mohammad et al. \citep{MohammadStanceandSentiment2016} and Lai et al. \citep{lai2016friends} exploited the new labels, showing that information on sentiment polarity and additional knowledge about the target of the opinion help in detecting stance.
In particular, Lai et al. \citep{lai2016friends} proposed an approach for detecting stance in Twitter that relies on domain knowledge by considering the context surrounding a target of interest. The proposed approach was evaluated on two targets of the SemEval2016-Task6: Hillary Clinton and Donald Trump. 
Three groups of features were considered: \emph{structural} (hashtags, mentions, punctuation marks, etc.), \emph{sentiment} (a set of four lexica to cover different facets of affect ranging from prior polarity of words such as AFINN \cite{Nielsen2011} and Hu\&Liu Lexicon, to fine-grained emotional information such as LIWC \cite{Pennebaker01} and the Dictionary of Affect in Language was exploited), and \emph{context-based} features.
Attempting to capture the contextual information surrounding a given target, the authors use the concepts ``friends'' and ``enemies'' for defining the relationships between the target and the politicians and the parties mentioned in the text.
The proposed approach outperform the state-of-the-art results, showing that information about enemies and friends of politicians help in detecting stance towards them.
In Lai et al. \cite{LaiExtractingGraph2017} the role of social relations was been analyzed together with the users' stance towards the BREXIT referendum.
The study shows two main results that may be of particular interest for addressing SD: users sharing the same stance towards a particular issue tend to belong to the same social network community, and users' stance diachronically evolves.
%Here, using the same dataset, we deeply investigate the linguistics aspects of the SD analysis.
A similar experiment has been performed by Lai et al.  \cite{Lai2018-StanceEvolution,LAI2019LKE} analyzing the political debate on Twitter about the Italian Constitutional referendum held in 2016.
Notice that the importance of using contextual features has been highlighted by many scholars in relation also with different NLP tasks aimed at detecting subjective information in user-generated contents. For instance,  
Wallace et al. \citep{wallace2015sparse} used communities belonging to different Reddit's subreddits in order to improve irony detection. Furthermore, Bamman and Smith \citep{bamman2015contextualized} used information such as historical terms, topics, and sentiment from both author and audience in order to improve sarcasm detection. 
%Lai et al. \cite{lai2016friends}, instead, used common knowledge information about the Party Presidential Primaries for the Democratic and Republican parties in order to improve SD.
The current proposal is settled on this line of research: we aim at exploring the importance of information about interpersonal relations, time evolution, and common knowledge about the target in SD, which can be all seen as different facets of context.

Recently, stance detection was addressed also in the context of political debates featured by languages different than English. The very controversial issue {\em Independence of Catalonia} was chosen as target in order to perform Stance Detection in Tweets 
(StanceCat\footnote{Detecting the gender of the author of a given tweet was also a sub-task to be addressed in the shared task.}) in the framework of the evaluation campaign IberEval 2017 \cite{overviewStanceCat2017}. 
The aim of the shared task was to detect the author's stance towards the Independence of Catalonia in tweets written in Spanish or Catalan, collected during the regional elections in Catalonia, which took place on September 27, 2015. 
A dataset in both languages was released, containing a set of tweets manually annotated as in favor, against or neutral towards the target of interest.
Well-known classifiers such as SVM and novel techniques such as deep learning approaches were used by the ten different teams participating in the shared task.
As features n-grams and word embeddings were the most used.
The best performed approach in both languages on the shared task, called iTACOS \cite{lai2017itacos}, consisted in a supervised approach that considers three groups of features: \emph{Stylistic} (bag of: n-grams, char-grams, part-of-speech labels, and lemmas), \emph{Structural} (hashtags, mentions, uppercase characters, punctuation marks, and the length of the tweet), and \emph{Context} (the language of each tweet and information coming from the URL in each tweet). The obtained results validate the importance of considering contextual information in stance detection tasks.

The political debate around the BREXIT referendum has been at the center of a growing interest, before and after the outcome of the vote. The kind of shock related to the outcome of the Brexit referendum is leading many scholars in different disciplines to focus on the stance dynamics underlying this political debate, with the aim to understand the processes that led to the result. Social media data could contribute also to shed some light on such dynamics related to political sentiment \citep{messinafersinireport2017}. An annotated corpus of tweets and news blogs about the BREXIT Referendum, collected in the month preceding the referendum date (June 23rd 2016), has been discussed in \cite{celli-EtAl:2016:PEOPLES}. 
It includes sentiment and agreement/disagreement labels on isolated posts. Nevertheless, the goal of that study is different, mainly targeted at forecasting the referendum outcome.
Our main aim here is, instead, to investigate the the predictive power of community detection features in SD and the 
importance of taking a diachronic perspective on stance, that, at the best of our knowledge, has been never studied before within this research community. 

\subsubsection*{Community detection and stance} 
Social media foster novel possibilities to investigate social networks embedding new forms of interpersonal relations. Research in {\em Network Science} have, thus, focused on exploring social networks, with the twofold aim to corroborating sociological theories and providing evidences to develop new ones \cite{gonccalves2011modeling,lazer2009life,gonzalez2008understanding,weng2015attention}. 

In this work we exploited a classical method to analyze the network structure, which is commonly used in social media networks, Community Detection (CD), that consists in identifying groups or communities in a given large scale network  \cite{fortunato2010community}.
Some recent works attempted to investigate the social media network structure in relation with sentiment information extracted from posted contents.
Xu et al. \citep{Xu:2011:SCD:1940761.1940913} introduce the concept of {\em sentiment community},
trying to maximize both the intra-connections of nodes and the sentiment polarities using ratings of movies information collected from Flixster\footnote{\url{http://www.flixster.com}}. Deitrick et al. \citep{deitrick2013mutually} combined SA and CD techniques on Twitter by using replies, mentions and retweets, hashtags, and sentiment classification of tweets to iterative increment edge weights in a social networks based on follower and friend relations.
Those works showed that CD and SA can be mutually supportive. 
However,
to the best of our knowledge, community detection techniques have been never used for detecting stance in Twitter. Some preliminary results in this direction have been reported in \cite{LaiExtractingGraph2017} that investigate the BREXIT debate for inspecting users' social network based on followings relations. This work, using the TW-BREXIT corpus for automatically predicting the stance of the increased number of unannotated users that took place in the debate on Twitter, showed that users having the same stance towards this particular issue tend to belong to the same social network community and that the neighbours are more likely to have similar opinions. In this paper, we present an enhanced stance model, enriched with the novel set of contextual features described in details in the next section, and present an in-depth evaluation of the model in a set of stance classification experiments on the TW-BREXIT corpus.

\section{Methodology}
\label{se:features}

Our methodology comprehends two steps. First, we developed a novel annotated corpus for studying stance in a diachronic perspective, which takes into account evolution of users' stance over time. Then, a novel set of features related to community and diachronic evolution has been introduced and evaluated in a set of stance classification experiments. 
The development of the Twitter annotated corpus TW-BREXIT, with focus on the debate on the BREXIT referendum which took place in UK, is described in the next section. 
Here, we will quickly describe our features, with a focus on the rationale behind the choice of the set
of novel features introduced to capture information related to community, diachronic and contextual facets of stance, which were also influencing the design of a novel annotation scheme. 

\subsection{Context-based features}

\subsubsection{Community features}
According to the homophily principle observed in on line social media as well as traditional social network \cite{mcpherson2001birds} people tend to bond with similar persons. Therefore, an interesting hypothesis to test in polarized political debate contexts is if people with the same stance belong to the same social media network community \cite{CalaisGuerra:2011:BOT:2020408.2020438,ICWSM112847}.
In particular, we assumed that the value homophily is involved \cite{Lazarsfeldfriendship1954}, considering that Twitter users
tend to bond with others who think in a similar ways, regardless of any differences in their status characteristics (i.e. gender,
age, social status).
Intuitively, the feature is based on the creation of a social network in witch a relation between two Twitter's users exits if one {\em follows} the other (in Twitter the term ``to follow'' refers to the specific relationship a user can create to be a follower of another user). After extracting the social communities from the network structure, the process results into the definition of the {\em community-context-based} feature.
It consists in a binary feature vector including one element for each detected community; value is 1 when the community corresponds to the one the tweeter belongs to, 0 otherwise.
%the element corresponding to the social community the tweeter belongs to is one.

\subsubsection{Diachronic evolution features}
People's opinion is influenced not only by pre-existent ideology and party identification, but also by information about events happened during the political discussion \cite{gelman1993american}.
Therefore, we hypothesize that the evolution of the political debate affects the stance of each voter. 
Dividing the dataset of Tweets in discrete temporal phases delimited by significant events occurred around the consultation period, could allow to track the evolution of stance of users involved in the online debate.
This assumption does not necessarily imply that users effectively change opinion, but that something
changes in the way they write about the topic.
The feature {\em diachronic-evolution-context-based} 
consists in a binary feature vector composed of an element for each considered time window.
The value of the element corresponding to the time window in which the user posts the tweet is set to 1. 
%is included in our SD model, taking into account three temporal frames: ``Referendum Day'', ``Outcome Day'', and ``After Pound Falls''.

\subsubsection{Common knowledge features}
There is a general agreement on the idea that language cannot be investigated in isolation from culture and social organization.
%of the investigated society. 
Such elements define an important context surrounding the event to be examined, and the use of resources embedding external knowledge can be important for a better interpretation \cite{duranti1992rethinking}.
%we first create a gazetteer of parties and politicians involved in BREXIT using both the Wikipedia and DBpedia resources (details on the entities extraction process in Section~\ref{CommonKnowledgeretrieving}).
The external knowledge could be extracted from resources nowadays available such as Wikipedia and DBpedia.
To address this issue, we introduced two binary features considering the relations of friendship and enmity among the target of interest and the related entities such as politicians and parties mentioned in the text of the tweet as a signal of stance that should be taken into account:
{\em party-stance-context-based}, a binary sub-feature vector of three elements  (party\_against, party\_favour, party\_neutral) considering the presence of a mentioned party in the text and its corresponding stance;
{\em politician-stance-context-based}: a binary feature vector of three elements (politician\_against, politician\_favour, politician\_neutral) considering the presence of a mentioned politician in the text and her corresponding stance.
Moreover, we introduce another binary feature of two elements, {\em explicit-stance-context-based}, considering the words used for expressing the stance (in the context of BREXIT the words are ``remain'' and ``leave'').
Aggregating the three sub-features, the \textit{Common knowledge features} feature consists is in a binary feature vector of seven elements.

\subsection{Sentiment-based features}
Stance detection is strongly related to sentiment \cite{lai2016friends,mohammad-EtAl:2016:SemEval,MohammadStanceandSentiment2016,sobhani-mohammad-kiritchenko:2016:interactionsentiment,zhang-lan:2016:SemEval}. 
We are not aware of sentiment analysis lexica retrieved specifically in the political domain; thus, we exploited a wide range of resources available for English \cite{SemAspinSA17}. We used a set of four lexica to cover different facets of affect, ranging from sentiment polarity of words to fine-grained emotional information \cite{lai2016friends}: AFINN \cite{Nielsen2011ANEW}, Hu\&Liu  \cite{Hu:2004:MSC:1014052.1014073}, LIWC  \cite{Pennebaker:01}, and DAL  \cite{Whissell01102009}.
AFINN was selected since contains several slang and profanity;  Hu\&Liu and LIWC, since they are widely used in tasks related to analysis of subjective information, and DAL in order to explore different emotional dimensions.
Therefore, the \textit{Sentiment-based} feature consists in a continuous feature vector of six element. Three elements respectively store the sum of the polarity calculated by AFINN, Hu\&Liu, and LIWC (considering as 1 the presence of a positive word in the text and as -1 the presence of a negative one) and other three elements for storing the the sum of the values of each word for each of the three emotional dimensions explored by DAL: pleasantness, activation, and imagery.

\subsection{Structural features}
We also experimented structural characteristics of tweets taking into account the use of metadata and punctuation marks \cite{evans2016}.
Therefore, the \textit{structural-based} features consists in a continuous feature vector of nine elements containing the number of elements of the following structural characteristics extracted from the text: number of hashtags, number of mentions, and number of punctuation marks (i.e., frequency of exclamation marks, question marks, periods, commas, semicolons, and, finally, the sum of all the punctuation marks mentioned before). In addition, binary unigrams of bag of hashtags and of bag of mentions are also included.

\section{The TW-BREXIT corpus}
\label{se:dataset}

We analyzed the political discussion in United Kingdom about the European Union membership referendum commonly known as BREXIT, an abbreviation
for ``British exit''. 
%The referendum that took place on Thursday 23 June 2016 in the United Kingdom and Gibraltar to gauge support for the country's continued membership in the European Union. 
We gather about 5M English tweets containing the hashtag \#brexit using the Twitter Stream API in the time-frame between 2016-06-22 and 2016-06-30 and we used the dataset in order to create a novel linguistic resource annotated for stance. A novel SD annotation schema has been applied to the data, focusing on users' stance in a diachronic perspective. 
First, we grouped tweets according to three 24-hour short and highly focused time steps forthcoming elections. 
For the ease of the reader, we defined each time windows by a name that corresponds to a relevant event that happened during the 24 hours: 
\begin{itemize}
\item ``Referendum Day'' (RD): includes the 24 hours preceding the polling stations closing. 
\item ``Outcome Day'' (OD): includes the 24 hours following the formalization of referendum outcome. 
\item ``After Pound Falls'' (APF): includes the 24 hours after the financial markets' turbulence occurred three day after the formalization of the referendum outcome.
\end{itemize}
%
%\begin{itemize}
%\item ``Referendum Day'': includes the 24 hours preceding the polling stations closing.
%\item``Outcome Day'': includes the 24 hours following the formalization of referendum outcome.
%\item``After Pound Falls'': includes the 24 hours after the financial markets' turbulence occurred  
%three day after the formalization of the referendum outcome.
%\footnote{The financial markets badly %reacted to the referendum outcome in %days 25th and 27the of June.}
%\end{itemize}
%
%In Figure \ref{characteristicsoftemporalphases} a sketch of the three phases is given.

%\begin{figure}[t]
%\includegraphics{}
%\caption{Figure caption.}\label{f1}
%\end{figure}
\iffalse
\begin{figure}[!htb]
\begin{center}
\includegraphics[width=50mm,scale=0.5]{TIME1.png}
\caption{Selected clear-cut events related to the referendum and defined time windows}\label{characteristicsoftemporalphases}
\end{center}
\end{figure}
\fi

Then, we randomly selected a sample of  600 users over 5,148 that wrote at least 3 tweets in each temporal interval. We decided to require three tweets for each user in each time windows due to our assumption is that human annotators may guess more easily the user's stance considering a context of several tweets instead of only one.
Therefore, we defined a {\em triplet} as a collection of three random
tweets written by the same user in a given time interval.
Finally, we created the TW-BREXIT corpus, which consists of 1,800 triplets. Overall, for each of the selected 600 users, we have three triplets in the corpus, one for each time interval. %This corresponds to have a total of 5,400 annotated tweets. 
%in the corpus.

\subsection{Annotation Process}
We employed CrowdFlower\footnote{\url{http://www.crowdflower.com/}}, a popular crowd-sourcing
resource, to annotate the corpus. Eligible annotators (i.e. ``contributors'') were required to live in the UK or in Gibraltar, since we wanted to be sure to they were directly involved in the political debate at issue, and aware about the local political situation. 
The proposed HIT (Human Intelligence Tasks\footnote{This is the expression for denoting questionnaires posted on a Crowdflower's job.}) request to the human contributors to annotate the user's stance on the target BREXIT (i.e. UK exit from EU).
In particular, given a triplet posted by a user, they had to infer the user's stance.
The instructions given to contributors for determining stance are shown below.
\vspace{0.3cm}
\hrule
\vspace{0.3cm}

{\em ``What is the stance of the user that wrote those three messages?''}

The available answers are: 
\begin{itemize}
\item \textbf{Leave}: you think that the user is {\em in favor} of the UK exit from EU.
\item \textbf{Remain}: you think that the user supports staying within the EU, being {\em against} the UK exit from EU.
\item \textbf{None}: you could not infer user's stance on BREXIT due to:
\begin{itemize}
\item all the messages are unintelligible
\item the user does not express express any opinion about the target
\item the user expresses opinion about the target, but the stance is unclear.
\end{itemize}
\end{itemize}

%Note: people that wrote messages are not necessarily from U.K., but they can express their opinion anyway.

\vspace{0.3cm}
\hrule
\vspace{0.3cm}

It is worth to be noticed that the classification task is very similar to the stance classification task formulated in Semeval-2016 \cite{mohammad-EtAl:2016:SemEval}. Indeed, the labels ``favor'', ``against'', and ``neither'' can be considered equivalent to ``leave'', ``remain'', and ``none'', respectively, if used to define stance towards the target ``UK exit from EU'' (BREXIT). 
Similarly to Mohammad et al. \cite{mohammad-EtAl:2016:SemEval}, we use only a label ``none'' for identifying both cases when the annotator can infer from the tweet that the tweeter has a neutral stance towards and when there is no clue in the tweet to reveal the stance of the tweeter towards the target.

We required two judgments for each triplet. When the two contributors disagree on the stance evaluation of a triplet, we requested an additional judgment by a new contributor.
Crowdflower provides a quality control mechanism to evaluate contributor reliability based on answers given to a set of test questions. Two human domain experts created 70 balanced test questions w.r.t. the three stance labels. We set the reliability threshold for our job, requiring that contributors should have correctly answered to at least 80\% of the test questions proposed during the task. %Annotation guidelines were prepared and provided to contributors\footnote{ Now available to reviewers.}.
%a public repository\footnote{ \sloppy\href{https://github.com/foranonymoussubmissions/submission/tree/master/data/annotated\%20\%20corpus}{github.com/foranonymoussubmissions/submission}}.

\subsubsection{Annotation Results}
The trusted contributors that perform the annotation process were twenty-nine\footnote{Four contributors did not pass the quality control during the annotation task and they were considered untrusted.}.
Crowdflower provides a measure of the
agreement between contributors calculated as the average confidence obtained by each HIT, which results to be 91.04\%.
%\footnote{
%2\/2*1356+2\/3*404+1\/3*40. 
%We also calculated the inter-annotator %agreement (IAA) for the question (i.e. %the percentage of times two annotators %agree with each other) according to %the method exposed in \cite{Mohammad:2015:SEP:2793724.2793929}, %which is 65.48\%.}.

In order to select the true label we used majority voting.
Overall, the final gold TW-BREXIT resulted in 1,760 labeled triplets, after removing triplets in
disagreement (where all three contributors provided different annotations). The corpus and the annotation guidelines are available 
to the community\footnote{\url{https://github.com/mirkolai/leave-or-remain}}.
 %\sloppy\href{https://github.com/foranonymoussubmissions/submission/tree/master/data/annotated\%20\%20corpus}{github.com/foranonymoussubmissions/submission}
%
%Table~\ref{labeldistibution} shows the label distribution.
\begin{table}[ht]
\centering
\small
\begin{tabular}{lll}
\hline
\textbf{Leave} & \textbf{Remain} & \textbf{ None} \\ \hline
961 (51\%)    & 236 (14\%)     & 563 (35\%)              \\ \hline
\end{tabular}
\caption{\label{labeldistibution}Label distribution}
\end{table}
The unbalanced distribution for stance is not unexpected. Indeed, we used the hashtag \#brexit for quickly collecting data, due to its wide use in the debate for marking posts which express different stance on the target. However, also apparently neutral hashtags are often biased, and a recent study \cite{howard2016bots} shows that most of the tweets containing \#brexit were posted from people that express stance in favor of BREXIT. The bias is not critical here, since our focus is on stance detection and evolution, and not in predicting the referendum outcome.
%So no wonder that an unbalanced %distribution was obtained.

\begin{figure}[!htb]
\begin{center}
\includegraphics[width=2.7in]{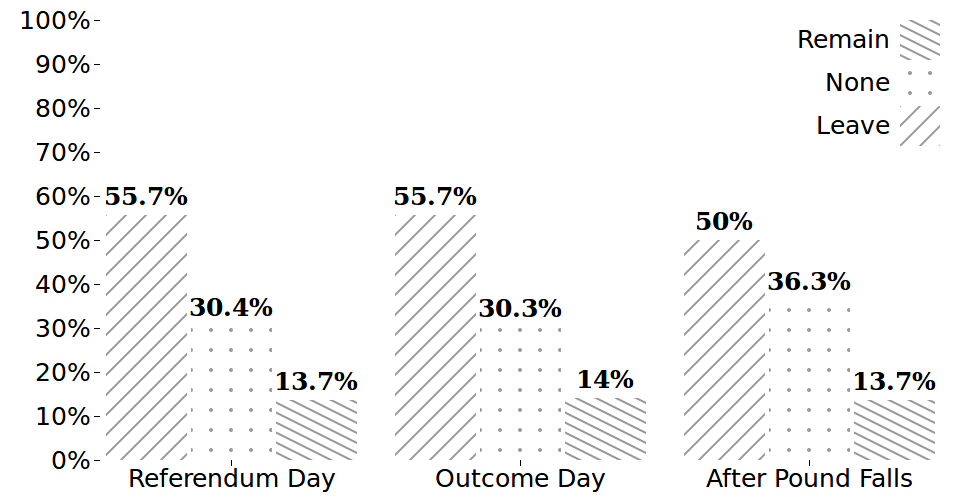}
\caption{Label distribution over the time}
\label{Label-distribution-over-the-time}
\end{center}
\end{figure}

Most interestingly, the label distribution changes over the time as shown in Figure~\ref{Label-distribution-over-the-time}.

% \iffalse
% It is important to notice that label distribution changes over the time as Figure \ref{Label-distribution-over-the-time} shows.

% \begin{table}[ht]
% \centering
% \small
% \begin{tabular}{c|c|c|c|}
% \cline{2-4}
%                                                                                   & \multicolumn{1}{l|}{\textbf{Leave}} & \multicolumn{1}{l|}{\textbf{Remain}} & \multicolumn{1}{l|}{\textbf{ None}} \\ \hline
% \multicolumn{1}{|c|}{\begin{tabular}[c]{@{}c@{}}Referendum \\ day\end{tabular}}    & 55.67\%                             & 13.67\%                              & 30.67\%                                         \\ \hline
% \multicolumn{1}{|c|}{\begin{tabular}[c]{@{}c@{}}Outcome \\ day\end{tabular}}       & 55.67\%                             & 14\%                                 & 30.33\%                                      \\ \hline
% \multicolumn{1}{|c|}{\begin{tabular}[c]{@{}c@{}}After Pound\\  falls\end{tabular}} & 50\%                                & 13.67\%                              & 36.33\%                                      \\ \hline
% \end{tabular}
% \caption{\label{Label-distribution-over-the-time}Label distribution over the time}
% \end{table}
% \fi

The label distribution changes consistently in the ``After Pound Falls" temporal interval, when we observe a drop in the percentage of label ``leave'', and an increase for what concerns the label ``none''.

We also explored if users’ stance changes over time. We find that 57.66\% of the users was labeled with the same stance in all three intervals (37.16\% leave, 15.5\% none, 5\% remain).
Very interestingly, 42.33\% of users' stance changes across different time intervals. In particular, 9.5\% of users' stance varies from ``leave'' to ``none'' (7\% leave $\,\to\,$ leave   $\,\to\,$  none; 2.5\%   leave $\,\to\,$  none $\,\to\,$  none).
Overall, the analysis of the corpus supports the conjecture that there is a relation between diachronic evolution and stance.
Furthermore, we observed a progressive decrease of the agreement (93.72 in the
RD interval; 90.61 in the next OD interval; 88.78 in the last APF interval).
The analysis of the manual annotated corpus could imply that something changed in users’ communication style up to the
point that annotators used a different label to annotate the stance of the same user in different time windows. Furthermore,
the progressive decreasing of the agreement among annotators could mean that users tend to express their opinion in a less
explicit way in the following phases of the debate. Also Messina et al. \cite{messinafersinireport2017} show how people’s views over technical
arguments about BREXIT could quickly change due to emotive events (i.e. national humiliation). For this reason we are
very encouraged by this preliminary analysis to keep trying to investigate stance in a diachronic perspective.

\section{Experiments and Results}
\label{se:experiments}

In this section, we first describe the methods used for detecting communities in the social network and for gathering common knowledge. 
Then, we report on three types of classification experiments for SD on the TW-BREXIT corpus: stance at the triplet level, at tweet level, and at tweet level in different temporal intervals. 
The source code of the experiments are available for the community\footnote{\url{https://github.com/mirkolai/leave-or-remain}}.
%in a public repository: \sloppy\href{https://github.com/foranonymoussubmissions/submission}{github.com/foranonymoussubmissions/submission}.

\subsection{Retrieving common knowledge}
\label{CommonKnowledgeretrieving}
For retrieving common knowledge, we created a gazetteer of relations (politician$\,\to\,$stance and party$\,\to\,$stance) using sources freely available such as Wikipedia and DBpedia. 
First, We extracted the declared stance (leave, remain, none) of UK's, Northern Ireland's, and Gibraltar's parties from Wikipedia~
\footnote{\url{https://en.wikipedia.org/wiki/United_Kingdom_European_Union_membership_referendum,_2016}}. 
Second, we gathered political affiliation of each politician in DBpedia, and we inferred politicians' stance from the stance of the party they are affiliated to.
We populated the gazetteer with politicians' and parties' alias provides by DBpedia and the inferred stances.
%We used the alias of named entities of parties and politicians, and inferred entity's stance only from extracted relations.
%Deficits of our approach are mainly %three: first, we are not able to %evaluate NER of politicians and parties based on created gazetteer.
%Second, 
%we found two or more politicians for %only one mention in a case of %homonymous (there are no cases of %parties' homonymous). 
%Third, 
One problem we had to face is that our knowledge sources are not constantly updated. Several politicians changed their political affiliation during their career and we couldn't infer the current political affiliation.
%(e.g., DBpedia provided historical information in alphabetical order not in chronological order). 
To deal with such cases, if a politician was affiliated at party with a  different stance in the past, we infer several stances for her.
Overall, we automatically obtained 225 alias for 17 parties and 5945 alias for 1838 politicians (48 of them having multiple stances due to the reasons highlighted above).
The feature extraction consists in giving the values of the elements of the feature vector \textit{Common knowledge features}. For example, the element party\_against is set to one if the text contains a party identified to have campaigned for a ``remain'' vote. 
%they have different political %affiliation in their political career).

\subsection{Community detection process}
\label{community_detection_process}
\begin{figure}[!htb]
\begin{center}
\includegraphics[width=2.8in]{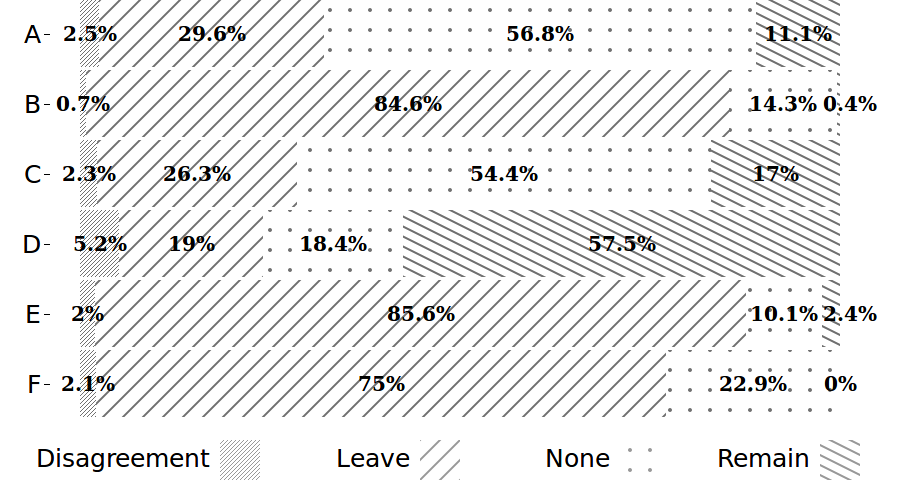}
\caption{\label{Users-stance-distribution-over-communities}
The figure shows the users' stance distribution of users belonging to each community. Only the 600 users resulting from the manual annotation are considered. The distribution is expressed as the average over the three temporal phases.}
\end{center}
\end{figure}

First, we used Twitter API ``GET friends/id'' in order to gather the list of the {\em friends} (the users that a user follows) of 5,148 users who posted at least three tweets containing the hashtag ``brexit'' in all the defined temporal intervals.
Then, we created a graph that consisted of about 13M edges and 4M users and friends.
We filtered a sub-graph consisting of about 200K nodes after removing the friends have less than 10 relations (we consistently reduced graph dimension).
We decided to use the Louvain Modularity algorithm\footnote{We used the software package NetworkX.} in order to extract social media network communities, since it performs better in terms of computer time and modularity compared to other methods \cite{Blondel2008fast}. 
The algorithm extracts 6 communities; we added a seventh community for 195 users that were isolated from the graph after the filtering.
Figure~\ref{Users-stance-distribution-over-communities} shows the average of the distribution over the communities of the 600 users' stance resulting from the manual annotation process. 

We observed that the percentage of users' stance in community D is evidently biased towards the stance ``remain'' (about 57\% users were annotated with the label ``remain''); in communities B, E, and F towards the stance ``leave''(more then 70\% users with the label ``leave''); in communities A and C towards the stance ``none'' (more then 50\% users with the label ``none''). We also noted that
the disagreement among contributors is higher for the users belonging to the community D (the annotators do not reach an agreement for about more 5\% of cases), maybe because the hashtag \#brexit is biased in favor
of BREXIT \cite{howard2016bots} and might have contributed to create more ambiguity during the annotation process in a community mostly composed by users annotated with the stance ``remain''.

\subsection{Classification experiments}

We experimented the use of several supervised learning algorithms - Na{\"i}ve Bayes (NB), linear support vector machine (SVM), Random Forest (RF), Decision Trees (DT) - on the TW-BREXIT\footnote{We used the scikit-learn (\url{http://scikit-learn.org}) implementation of the machine learning algorithms with default parameters.}.
In addition, we experimented with different feature sets for SD, and evaluated them 
%with the average on 10 different 
performing 5-fold cross validation for each run.
We used the macro-average of the AGAINST and FAVOR F1-score metrics and the baselines such as Majority Class, SVM-unigrams (word unigram features), SVM-ngrams (1-, 2-, and 3- word grams features and  2-, 3-, 4-, and 5- character grams features) proposed in Semeval-2016 \cite{mohammad-EtAl:2016:SemEval}.
The macro-average of the F1-score metrics was redefined, replacing labels ``favor'' and ``against'' with labels ``leave'' and ``remain'', respectively:

$${ F\textsubscript{avg} =  \frac{F_{leave}+F_{remain}}{2} }$$

We carried out three kinds of experiments:
\begin{itemize}
\item \textbf{Stance at the triplet level}: we looked for the better feature combination set in order to predict users' stance extracting features from the textual contents of the three tweets grouped in a triplet.
\item \textbf{Stance at the tweet level}: we tried to find the best feature combination set in order to predict users' stance extracting features from a tweet isolated from a triplet inheriting triplet label. 
\item \textbf{Stance at the tweet level in different temporal intervals}: we used the best feature combination obtained in the experiment ``Stance at the tweet level'', by grouping single tweets from each triplet according to the temporal interval.
\end{itemize}

\subsubsection{Stance at the triplet level}    
First, we experimented SD at the level of triplets. Here, the classifiers, similarly to the human annotators, are trained with triplets - three tweets for each user in each temporal phase.
In other words, the training and the test instances are triplets instead of single tweets.
Thus, we performed features extraction concatenating the textual content of three tweets (instead of relying only on the single tweet).
We experimented 63 different feature combination using six groups of features: 
\textit{BoW}, \textit{structural-based} (structural), \textit{sentiment-based} (sentiment), \textit{common-knowledge-context-based} (comm-know-cxt), \textit{diachronic-evolution-context-based} (de-cxt), \textit{community-context-based} (comm-cxt).
Results are showed in Table~\ref{exp3userlevel}.

\begin{table}[ht]
\centering
\small
\begin{tabular}{lcc}
\hline
Classifier & Feature set  & F\textsubscript{avg} \\ \hline
\textit{Baselines}\\ 
\hspace{0.3cm}MC & -  & 35.25  \\ 
\hspace{0.3cm}SVM  & unigrams  & 58.25  \\ 
\hspace{0.3cm}SVM  & ngrams  & 60.14 \\ 
\textit{Our Classifiers}\\ 
\hspace{0.3cm}NB  &BoW + comm-cxt                             & 53.77                                                    \\ 
\hspace{0.3cm}DT  & comm-cxt   & 63.74     \\ 
\hspace{0.3cm}RF  & comm-cxt   & 63.76     \\ 
\hspace{0.3cm}SVM             
&\begin{tabular}[c]{@{}c@{}}structural + sentiment +\\ de-cxt + comm-cxt\end{tabular} 
& \textbf{67.01}                                  \\ \hline
\end{tabular}
\caption{\label{exp3userlevel}Best feature set on stance at triplet level}
\end{table}

The \textit{BoW} and \textit{structural-based} features are relevant due to the presence of three tweets in a triplet (more words than in a text of a single tweet) respectively in Na{\"i}ve Bayes and SVM.
The \textit{community-context-based} feature is significant especially in Decision Tree and Random Forest.
In addition, all the best feature combinations for each classifier contain the \textit{community-context-based} feature. 
The \textit{diachronic-evolution-context-based} feature shows its relevance only in SVM.
Table \ref{Interesting-results-in-SVM-level-user} shows the results obtained in the ablation test using linear SVM, i.e. the machine learning algorithm that achieved the best performance in the above mentioned experiments\footnote{Notice that we will report on ablation test results only for classification experiments at the triplet level, since this is here our main novel focus for what concerns the task of detecting stance.}.
 $F_{avg}$ decreases of 14.6\% and 0.12\% removing singularly \textit{community-context-based}
and \textit{diachronic-evolution-context-based} features respectively. Removing only the \textit{common-knowledge-context-based} feature $F_{avg}$ improved of 0.49\%. Therefore, the
\textit{common-knowledge-context-based} feature does not improve $F_{avg}$ and the \textit{diachronic-evolution-context-based} feature is not decisive in results. Using the whole group of
\textit{context-based} features improves $F_{avg}$ more than using only the \textit{community-context-based} features (16.78\%).
%, the rank consider all 63 combination.
\begin{table}[ht]
\centering
\small
\begin{tabular}{lccc}
\toprule
Features   & F\textsubscript{avg} & Decreasing & \begin{tabular}[c]{@{}l@{}}Percentage\\ decreasing\end{tabular} \\ \midrule
All                 & 65.61                                   & 0                                       & 0\%                                                \\
All - context-based & 54.60                                   & -11.01                                  & -16.78\%                                           \\
All - comm-cxt      & 56.03                                   & -9.58                                   & -14.6\%                                            \\
All - de-cxt        & 65.53                                   & -0.08                                   & -0.12\%                                            \\
All - comm-know-cxt  & 65.93                                   & 0.32                                    & 0.49\%                                             \\
All - sentiment     & 65.99                                   & 0.38                                    & 0.58\%                                             \\
All - structural    & 65.81                                   & 0.2                                     & 0.3\%                                              \\
All - BoW           & 65.66                                   & 0.05                                    & 0.08\%                                             \\ \bottomrule
\end{tabular}
\caption{\label{Interesting-results-in-SVM-level-user}
Ablation Test (linear support SVM on stance at triplet level)}
\end{table}

Wallace et al. \citep{wallace2015sparse} already
noted how the mention of a specific political entity in a determined community could be a useful feature for irony detection
on political debates. \textit{Sentiment-based}, \textit{structural-based}, and \textit{BoW} features could be removed without significant influencing $F_{avg}$.

\subsubsection{Stance at the tweet level}
Second, we experimented SD at the level of tweet. The classifiers, differently from the situation faced by the human annotators, deal only with single tweets, isolated from triples, but inheriting the triplet label.
%We expected features based on BoW, %specially the SVM baseline, decrease %their results.
We experimented the same 63 different features combinations used in the previous experiment.
Table~\ref{exp1_bestfeatureset} shows the best feature set combination for each classifier.

\begin{table}[ht]
\centering
\small
\begin{tabular}{lccc}
\hline
Classifier & Feature set & F\textsubscript{avg}  \\ \hline
\textit{Baselines}\\ 
\hspace{0.3cm} MC   & -             & 35.25    \\ 
\hspace{0.3cm} SVM   & unigrams             & 51.98    \\ 
\hspace{0.3cm} SVM   & ngrams             & 52.66    \\ 

\textit{Our Classifiers}\\ 
\hspace{0.3cm}NB                                                                           &sentiment + community-ctx  
& 52.00                       \\ 
\hspace{0.3cm}DT    & community-ctx & 63.75    \\ 
\hspace{0.3cm}RF    & community-ctx & 63.74    \\ 
\hspace{0.3cm}SVM                                           
&\begin{tabular}[c]{@{}l@{}}BoW + sentiment +\\ de-cxt + comm-cxt \end{tabular}           & \textbf{65.68}                                     \\ \hline
\end{tabular}
\caption{\label{exp1_bestfeatureset}Best feature set on stance at tweet level}
\end{table}

%The SVM baselines are less than ones proposed in stance at tweet level, as expected (only one tweet instead of three implies less word in the vector representation). 
SVM obtains the best result using \textit{BoW}, \textit{sentiment-based}, \textit{diachronic-evolution-context-based}, \textit{community-context-based} based features. Furthermore, it is important to highlight that all classifiers improve consistently the Majority Class.
In addition, all the best feature combinations for each classifier contain the \textit{community-context-based} feature. In particular, the \textit{community-context-based} feature in Decision Tree and Random Forest singularly is the best feature. 
The feature \textit{diachronic-evolution-context-based} seems to be relevant only in SVM.
Moreover, Na{\"i}ve Bayes did not improve SVM unigram and ngram baselines.

\subsubsection{Stance in different temporal intervals}
We experimented with the best features sets obtained in the previous experiment over the three different temporal intervals selected.
We decided to carry out this experiment since we observed that the agreement between the annotators varies over the temporal intervals, in particular we observed a progressive decrease of the agreement (93.72 in the RD interval; 90.61 in the next OD interval; 88.78 in the last APF interval
%la media infatti è 91.037
).  We hypothesized that the proposed features could detect cues
which influence annotators’ agreement over the time windows. 
%in section \ref{sec:dataset}.
Table~\ref{exp2_bestfeatureset} shows the results obtained by the features set combinations over the three temporal phases. 

\begin{table}[ht]
\centering
\small
\begin{tabular}{lcccc}
\hline
Classifier  & Feature set  & RD & OD & APF \\ \hline
\textit{Baselines}\\ 
\hspace{0.3cm}MC & -                                                                    & 35.97                                                    & 36.07                                                 & 33.6                                                            \\ 
\hspace{0.3cm}SVM & unigrams  & 54.46 & 49.93 & 45.77 \\ 
\hspace{0.3cm}SVM & ngrams    & 56.60 & 48.49 & 45.35 \\ 
\textit{Our Classifiers}\\ 
\hspace{0.3cm}NB           
& \begin{tabular}[c]{@{}c@{}}sentiment +\\ comm-cxt\end{tabular}   & 47.56                                                    
& 54.20                                                 
& 43.25 \\ 
\hspace{0.3cm}DT  
& comm-cxt                                                  
& 67.07                                                    
& 61.95                                                 
& \textbf{62.22} \\ 
\hspace{0.3cm}RF
& comm-cxt                                                       & 67.14                                                    
& 61.96                                                 
& 62.07 \\ 
\hspace{0.3cm}SVM                                                      
& \begin{tabular}[c]{@{}c@{}}BoW  + \\sentiment +\\ comm-cxt\end{tabular} 
& \textbf{69.75}                                                  
& \textbf{62.03}                                               
& 58.30 \\ \hline
\end{tabular}
\caption{\label{exp2_bestfeatureset}Results in different temporal intervals}
\end{table}

We did not use the \textit{diachronic-evolution-context-based} feature, since the feature would become superfluous after grouping tweets by temporal interval. The $F_{avg}$ changes over the time for each classifier. We observe that $F_{avg}$ decreases for SVM and Na{\"i}ve Bayes in the time interval ``After Pound falls'' as with annotators' agreement. 
%This is in accordance with the manual %annotation agreement as well as with %SVM baselines.

\section{Conclusions}
\label{se:conclu}

In this work we investigated the use of several context-based features related to common knowledge, social network community, and diachronic evolution in the stance detection task. We focus on the political debate on BREXIT in Twitter. 
A novel annotation scheme, which takes into account the temporal evolution of stance has been proposed and applied to our social media data. 
Results of classification experiments confirm that the entire group of context-based features is very relevant for the stance detection task, in particular the community based feature. 
%The proposed group of features related to this aspect works very well with all tested learning algorithms.
Then, the analysis of the annotated corpus confirms that users' labeled stance may consistently change over temporal phases. We can only speculate that users not only could effectively change opinion, but also they could change their communication style, probably influencing annotators’ choices.
However, even if a deeper investigations on the possible causes of the opinion shifts are needed, calling also for competencies from other disciplines such as sociology or social psychology, this finding confirms that it is interesting to investigate SD in a diachronic perspective, since opinion fluctuations within the debates occur even in short time spans.
It also suggests that people's stance depends not only on their pre-existent ideology and party identification, but also on the information about events happened during the political discussion \cite{gelman1993american}. 

We are currently planning to investigate the debate about Brexit during the last three year particularly focusing on European elections in order to explore a wider time windows.

As a future work, a deeper linguistic analysis is also needed in order to clarify what has been observed here. In particular, we would like to investigate not only stance, but also the communications among unlike-minded users focusing on the use of rhetorical devices, such as sarcasm \cite{Sulis-2016Figurative}, and hostility. Here, we considered a static network structure, whereas on this future line of research it will be interesting to consider the {\em mutual evolution of stance and network structure} for observing the dynamics of the polarization within the debate.
Furthermore, the proposed method could be useful for training intelligent systems capable of  gathering and analyzing real time user generated contents from social media for supporting automatic prediction of citizens' stance toward public issues, based on big amount of people's opinions spontaneously expressed.
Thus, policy makers and public administrators could better meet population's needs and could have a data-driven support for developing policies to prevent strong polarization among different groups in society.

\section*{Acknowledgments}
The work of P. Rosso was partially funded by the Spanish MICINN under the
research projects MISMIS-FAKEnHATE on Misinformation and 
Miscommunication in social media: FAKE news and HATE 
speech(PGC2018-096212-B-C31) and PROMETEO/2019/121 (DeepPattern) of 
the Generalitat Valenciana. \\
The work of V. Patti and G. Ruffo was partially funded by Progetto di Ateneo/CSP 2016 {\em Immigrants, Hate and Prejudice in Social Media} (S1618\_L2\_BOSC\_01).
% So long and thanks for all the fish.

\bibliographystyle{plain}
\bibliography{main}

\end{document}